\documentclass[runningheads]{llncs}
\usepackage[T1]{fontenc}
\usepackage{graphicx}
\usepackage{hyperref}
\usepackage{color}
\usepackage[misc]{ifsym}

\usepackage{amssymb}
\usepackage{booktabs}
\usepackage{amsmath}
\usepackage{xspace}
\usepackage{mathtools}
\usepackage{subcaption}
\usepackage{wrapfig}
\usepackage{multicol}

\newcommand{\vspaceBelowLargeTable}[0]{\vspace{-4mm plus 0.5mm minus 0.5mm}}
\newcommand{\vspaceBelowLargeFigureCaption}[0]{\vspace{-6mm plus 0.5mm minus 0.5mm}}

\newcommand{\myParagraph}[1]{\textbf{#1.} \nolinebreak}

\newcommand{\method}[0]{DSV\xspace}
\newcommand{\methodLong}[0]{Discordance and Separability Validation\xspace}

\newcommand{\data}[0]{\mathbf{x}}
\newcommand{\augFunc}[0]{f_\mathrm{aug}}
\newcommand{\genFunc}[0]{f_\mathrm{gen}}

\newcommand{\trnData}[0]{\mathcal{D}_\mathrm{trn}}

\newcommand{\testData}[0]{\mathcal{D}_\mathrm{test}}

\newcommand{\detAll}[0]{\phi}
\newcommand{\detEnc}[0]{\phi_\mathrm{enc}}
\newcommand{\detDec}[0]{\phi_\mathrm{dec}}

\newcommand{\trnEmb}[0]{\mathcal{Z}_\mathrm{trn}}
\newcommand{\augEmb}[0]{\mathcal{Z}_\mathrm{aug}}
\newcommand{\testEmb}[0]{\mathcal{Z}_\mathrm{test}}
\newcommand{\testEmbN}[0]{\mathcal{Z}_\mathrm{test}^{(n)}}
\newcommand{\testEmbA}[0]{\mathcal{Z}_\mathrm{test}^{(a)}}

\newcommand{\trnPoint}[0]{\mathbf{z}_\mathrm{trn}}
\newcommand{\augPoint}[0]{\mathbf{z}_\mathrm{aug}}
\newcommand{\testPoint}[0]{\mathbf{z}_\mathrm{test}}

\newcommand{\testPointA}[0]{\mathbf{z}_\mathrm{test}^{(a)}}

\newcommand{\disVar}[0]{h_d}
\newcommand{\sepVar}[0]{h_s}

\newcommand{\setDist}[0]{d}

\newcommand{\finalLoss}[0]{\mathcal{L}_\mathrm{DSV}}
\newcommand{\alignLoss}[0]{\mathcal{L}_\mathrm{ali}}
\newcommand{\scaledLoss}[0]{\mathcal{L}_\mathrm{dis}}
\newcommand{\covLoss}[0]{\mathcal{L}_\mathrm{sep}}

\newtheorem{assumption}{Assumption}
\begin{document}
\title{DSV: An Alignment Validation Loss for Self-supervised Outlier Model Selection}
\titlerunning{DSV for Self-supervised Outlier Model Selection}
%
\author{Jaemin Yoo \and
Yue Zhao \and
Lingxiao Zhao \and
Leman Akoglu\textsuperscript{(\Letter)}}
\authorrunning{J. Yoo et al.}
%
\institute{Carnegie Mellon University, Pittsburgh, USA\\
\email{\{jaeminyoo, zhaoy,  lingxiao\}@cmu.edu, lakoglu@andrew.cmu.edu}}
\maketitle              
\begin{abstract}
Self-supervised learning (SSL) has proven effective in solving various problems by generating internal supervisory signals.
Unsupervised anomaly detection, which faces the high cost of obtaining true labels, is an area that can greatly benefit from SSL. 
However, recent literature suggests that tuning the hyperparameters (HP) of data augmentation functions is crucial to the success of SSL-based anomaly detection (SSAD), yet a systematic method for doing so remains unknown.
In this work, we propose \method (\methodLong), an unsupervised validation loss to select high-performing detection models with effective augmentation HPs.
\method captures the alignment between an augmentation function and the anomaly-generating mechanism with surrogate losses, which approximate the discordance and separability of test data, respectively.
As a result, the evaluation via \method leads to selecting an effective SSAD model exhibiting better alignment, which results in high detection accuracy.
We theoretically derive the degree of approximation conducted by the surrogate losses and empirically show that \method outperforms a wide range of baselines on 21 real-world tasks.

\end{abstract}
\section{Introduction}

Through the use of carefully annotated data, machine learning (ML) has demonstrated success in various applications. Nonetheless, the high cost of acquiring high-quality labeled data poses a huge challenge. 
A recent alternative, known as self-supervised learning (SSL), has emerged as a promising solution. 
Intuitively, SSL generates a form of internal supervisory signal from the data to solve a task, thereby transforming an unsupervised task into a supervised problem by producing (pseudo-)labeled examples. 
It has achieved remarkable progress in advancing
natural language processing \cite{baevski2022data2vec,elnaggar2021prottrans} and computer vision tasks \cite{chen2021empirical,Kolesnikov19Revisiting}.

SSL can be particularly advantageous when dealing with unsupervised problems such as anomaly detection (AD). 
The process of labeling for such problems, such as correctly identifying fraudulent transactions, can be challenging and expensive. 
As a result, a group of SSL-based AD (SSAD) approaches \cite{Bergman20GOAD,Golan18GEOM,Li21CutPaste} have been proposed recently, where the core idea is to inject self-generated pseudo anomalies into the training data to improve the separability between inliers and pseudo anomalies. 
To create such pseudo anomalies, one may transform inliers via data augmentation function(s) such as rotate, blur, mask, or CutPaste \cite{Li21CutPaste}, which are designed to create a systematic change without discarding important original properties such as texture or color depending on the dataset.

Despite the surge of SSAD methods, 
how to set the hyperparameters (HPs), e.g., rotation degrees, remain underexplored, which can significantly affect their performance \cite{yoo2022role}.
In the supervised ML community, these augmentation HPs are systematically integrated into the model selection problem to be chosen with a hold-out/validation set \cite{mackay2019self,zoph2020learning}. 
However, choosing the augmentation HPs has been arbitrary and/or ``'cherry-picked' in SSAD \cite{Bergman20GOAD,Golan18GEOM} due to the evaluation challenges.
Recent literature shows that the arbitrary choice of SSAD augmentation has implications \cite{yoo2022role}.
Firstly, due to the no-free-lunch theorem \cite{wolpert1997no}, different augmentation techniques perform better on different detection tasks, and arbitrary selection is thus insufficient. Secondly, in some cases, the arbitrary selection of augmentation HPs
can lead to a biased error distribution \cite{Ye21Bias}. 
Thus, augmentation HPs in SSAD should be chosen carefully and systematically.

\begin{figure}[t]
    \centering
    \begin{subfigure}{0.46\textwidth}
        \includegraphics[width=\textwidth]{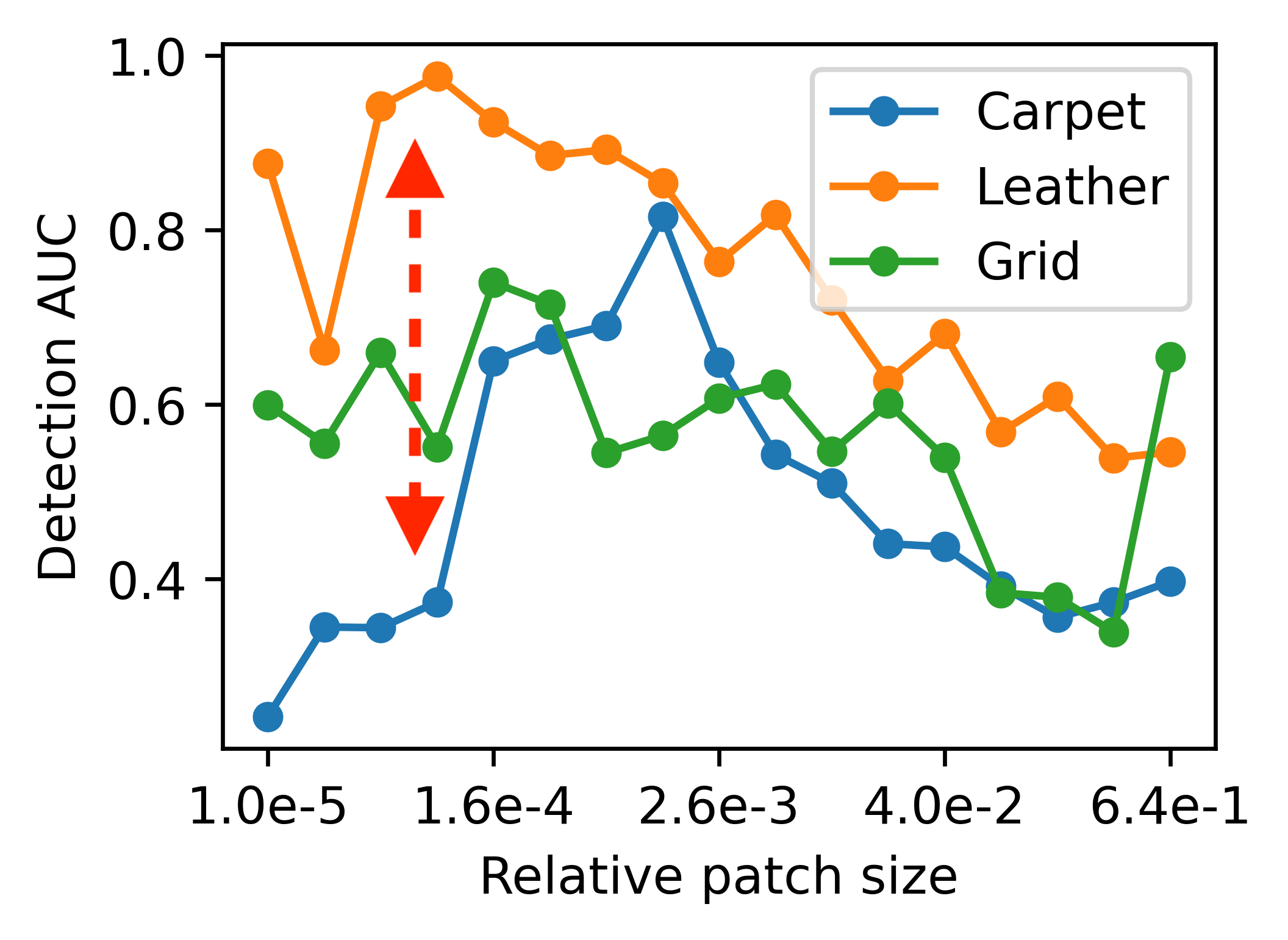}
        \caption{$\augFunc = \textrm{CutOut}$}
        \label{fig:motivation-1}
    \end{subfigure} \hspace{3mm}
    \begin{subfigure}{0.46\textwidth}
        \includegraphics[width=\textwidth]{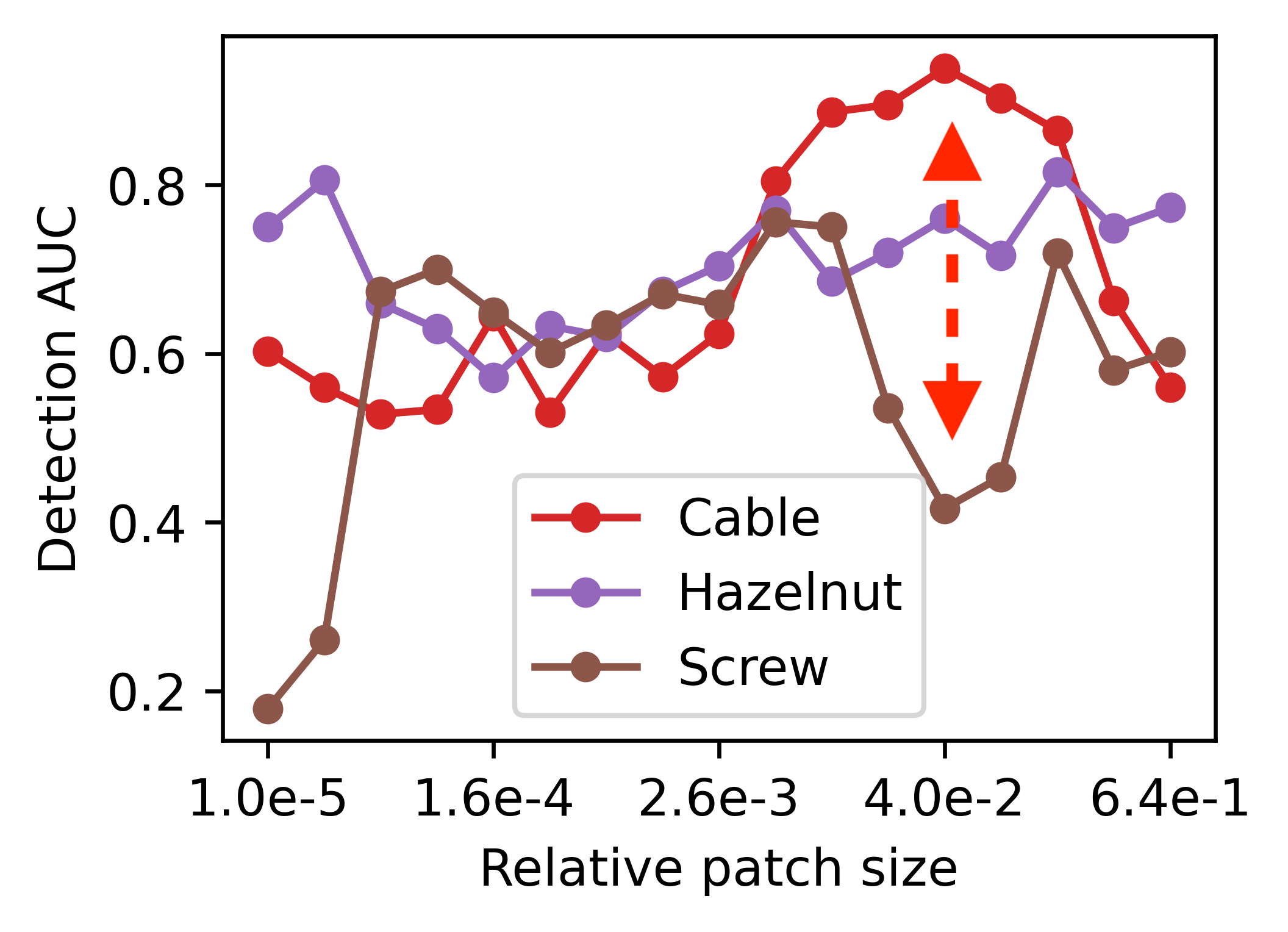}
        \caption{$\augFunc = \textrm{CutPaste}$}
        \label{fig:motivation-2}
    \end{subfigure}
    \caption{
        The performance of self-supervised anomaly detectors on the MVTec AD dataset with different hyperparameters of augmentation $\augFunc$.
        Each line is drawn from one of the 15 tasks that MVTec AD contains.
        The AUC changes from $0.242$ to $0.815$ based on the choice of hyperparameters (in Carpet), where the optimum depends on the type of $\augFunc$ and true anomalies.
    }
    \label{fig:motivation}
    \vspaceBelowLargeFigureCaption
\end{figure}

Fig. \ref{fig:motivation} shows how the performance of SSAD methods changes by the choice of augmentation HPs.
The CutOut \cite{Devries17CutOut} and CutPaste \cite{Li21CutPaste} augmentations are used for MVTec AD \cite{Bergmann19MVTecAD}, which is a real-world dataset for anomaly detection.
In Carpet of Fig. \ref{fig:motivation-1}, for example, the detection AUC changes from $0.242$ to $0.815$ with the choice of HPs.
The expected accuracy without prior knowledge is severely worse than its optimum, highlighting the importance of a proper HP choice, which is not even the same for different augmentation functions and tasks.

One solution is to select augmentation HPs in SSAD via unsupervised outlier model selection (UOMS) \cite{zhao2021automatic,zhao2022elect}, which aims to choose a good AD model and its HPs for a new dataset without using any labels. 
Given an underlying AD model, we may pair it with different augmentation HPs to construct candidate models to find the best performing one.
Existing UOMS approaches can be briefly split into two groups.
The first group solely depends on the model’s output or input data \cite{ma2023need}, while it cannot capitalize on the nature of SSAD. 
The second group uses learning-based approaches to select a model using the performances on (similar) historical datasets, while this prior information may be inaccessible.

In this work, we propose \method (\methodLong), an unsupervised objective function that enables the search for optimal augmentation HPs without requiring true labels.
The main idea of \method is to decompose the \emph{alignment} between data augmentation and true anomalies, which cannot be computed without labels but plays an essential role in estimating the detection performance, into \emph{discordance} and \emph{separability}.
Since each of them reflects only a part of the original alignment, the decomposition allows us to devise surrogate losses which effectively approximate the alignment in combination.

We summarize our key contributions below:
\begin{itemize}
    \item \textbf{Unsupervised validation loss for SSAD:}
        We propose \method, an unsupervised validation loss for the search of best augmentation HPs in SSAD.
        The minimization of \method leads to a high-performing AD model, which exhibits better alignment between augmentation and true anomalies.
    \item \textbf{Theoretical analysis:}
        We theoretically show that \method is an effective approximation of the alignment between data augmentation and true anomalies, and its minimization leads to well-aligned augmentation HPs.
    \item \textbf{Extensive experiments:}
        We conduct extensive experiments on 21 different real-world tasks.
        \method surpasses 8 baseline approaches, showing up to $12.2\%$ higher average AUC than the simple average.
        We also perform diverse types of ablation and case studies to better understand the success of \method.
\end{itemize}

\noindent \myParagraph{Reproducibility}
All of our implementation and datasets are publicly available at \url{https://github.com/jaeminyoo/DSV}.

\section{Problem Definition and Related Works}

\subsection{Problem Definition}

Let $\augFunc: \mathbb{R}^m \rightarrow \mathbb{R}^m$ be a data augmentation function on $m$-dimensional data, such as the rotation of an image, which plays an important role in self-supervised anomaly detection (SSAD).
Then, we aim to solve the unsupervised outlier model selection (UOMS) problem, focusing on the hyperparameters (HP) of $\augFunc$, based on observations that choosing good HPs of $\augFunc$ is as important as selecting the detector model or $\augFunc$ itself.
We formally define the problem as Problem \ref{def:problem}.

\begin{problem}
    Let $\trnData$ be a set of normal data, and $\testData$ be an unlabeled test set containing both normal data and anomalies.
    Given $\trnData$, $\testData$, and a set $\{ \phi_i \}_i$ of detector models, each of which is trained on $\trnData$ with an augmentation function $\augFunc$ of different hyperparameters, our goal is to find the model $\phi^*$ that produces the highest detection accuracy on $\testData$, without having true labels.
\label{def:problem}
\end{problem}

We also assume that every detector model $\detAll = \detEnc \circ \detDec$ which we consider for UOMS consists of an encoder $\detEnc \in \mathbb{R}^m \rightarrow \mathbb{R}^l$ and a decoder $\detDec \in \mathbb{R}^l \rightarrow \mathbb{R}$, which is typical of most AD models based on deep neural networks.

\subsection{Self-supervised Anomaly Detection (SSAD)}

With the recent advance in self-supervised learning, SSAD has been widely studied as a promising alternative to unsupervised AD models.
The main idea is to create pseudo-anomalies and inject them into a training set, which contains only normal data, to utilize supervised training schemes.
For example, a popular way is to learn a binary classifier that divides normal and augmented data \cite{Li21CutPaste} or an $n$-way classifier that predicts the type of augmentation used \cite{Bergman20GOAD,Golan18GEOM}.
Many SSAD methods have shown a great performance on real-world tasks \cite{Qiu21NeuTraL,Sehwag21SSD,Shenkar22ICL,Sohn21DeepOC}.

However, most existing works on SSAD are based on arbitrary and/or cherry-picked choices of an augmentation function and its HPs.
This is because AD does not contain a labeled validation set for a systematic HP search unlike in typical supervised learning.
A recent work \cite{yoo2022role} pointed out such a limitation of existing works and showed that augmentation HPs, as well as the augmentation function itself, work as important hyperparameters that largely affect the performance on each task.
Thus, a systematic approach for unsupervised HP search is essential to design generalizable and reproducible approaches for SSAD.

\subsection{Unsupervised Outlier Model Selection (UOMS)}
\label{subsec:uoms}

UOMS aims to select an effective model without using any labels. Clearly, choosing the augmentation hyperparameters (HPs) of an AD algorithm in SSAD can be considered a UOMS problem.
In this case, a candidate model is defined as a pair of the underlying AD algorithm and augmentation HPs, and the goal is to choose the  one that would achieve high detection rate on test data.

Existing UOMS approaches can be categorized into two groups.
The first group uses internal performance measures (IPMs) that are based solely on the model's output and/or input data \cite{ma2023need}. 
We adopt three top-performing IPMs reported in \cite{ma2023need} as baselines (see \S \ref{subsec:baselines}).
The second group consists of meta-learning-based approaches \cite{zhao2021automatic,zhao2022elect}. In short, they facilitate model selection for a new unsupervised task by leveraging knowledge from similar historical tasks/datasets.
It is important to note that in this work we do not assume access to historical training data. Thus, learning-based UOMS approaches do not apply here.

\section{Proposed Method}

We introduce \method (\methodLong), our unsupervised validation loss for the search of augmentation HPs in SSAD.
The minimization of \method leads to better alignment between data augmentation and true anomalies, which in turn results in higher accuracy on anomaly detection.

\subsection{Definitions and Assumptions}

We first introduce definitions and assumptions on which \method is based.
We start by defining set distance and projection functions.
Note that by Definition \ref{def:set-distance}, the set distance $d$ satisfies the triangle inequality between three different sets.

\begin{definition}
    We define a set distance $d$ as the average of all pairwise distances: $\smash{d(\mathcal{A}, \mathcal{B}) = \frac{1}{|\mathcal{A}||\mathcal{B}|} \sum_{\mathbf{a} \in \mathcal{A}} \sum_{\mathbf{b} \in \mathcal{B}} \|\mathbf{a} - \mathbf{b}\|}$.
    We also represent the vector distance as $d$ for the brevity of notations: $d(\mathbf{a}, \mathbf{b}) \coloneqq d(\{\mathbf{a}\}, \{\mathbf{b}\})$.
\label{def:set-distance}
\end{definition}



\begin{definition}
    We define a projected norm as $\smash{\mathrm{proj}(\mathbf{a}, \mathbf{b}, \mathbf{c}) = \frac{(\mathbf{c} - \mathbf{a})^\top (\mathbf{b} - \mathbf{a})}{\|(\mathbf{b} - \mathbf{a})\|}}$.
    The meaning of $\mathrm{proj}$ is the norm of $\mathbf{c} - \mathbf{a}$ projected onto the direction of $\mathbf{b} - \mathbf{a}$, using $\mathbf{a}$ as the anchor point.
    Note that $\mathrm{proj}(\mathbf{a}, \mathbf{b}, \mathbf{c}) \leq \|\mathbf{c} - \mathbf{a}\|$.
\end{definition}

Then, we introduce an assumption on data embeddings.
Recall that our detector $\detAll = \detEnc \circ \detDec$ contains an encoder function $\detEnc \in \mathbb{R}^m \rightarrow \mathbb{R}^l$.
Let $\trnEmb$ and $\testEmb$ be sets of embeddings for training and test samples, respectively, such that $\trnEmb = \{ \detEnc(\data) \mid \data \in \trnData \}$ and $\testEmb = \{ \detEnc(\data) \mid \data \in \testData \}$.
Let $\smash{\testEmbN}$ and $\smash{\testEmbA}$ be the normal and anomalous data in $\testEmb$, respectively.
We also define $\augEmb = \{ \detEnc(\augFunc(\data)) \mid \data \in \trnData \}$ as a set of augmented embeddings.


\begin{assumption}
    By convention, we assume that training normal and test normal data are generated from the same underlying distribution.
    Let $\setDist(\trnEmb, \trnEmb) = \sigma$.
    Then, $\smash{\setDist(\testEmbN, \testEmbN) = \sigma}$ and $\smash{\setDist(\trnEmb, \testEmbN) = \sigma + \epsilon}$, where $\epsilon < \sigma$.
\label{assume:test-data}
\end{assumption}




\begin{figure}[t]
\centering
\includegraphics[width=0.85\textwidth, trim={0.5cm 0 2.1cm 0}, clip]{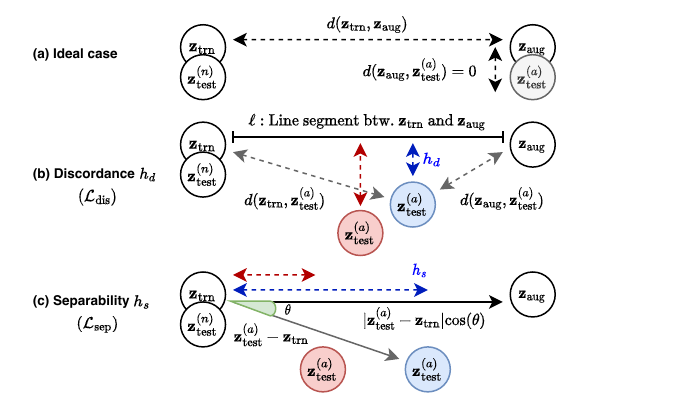}    
\caption{
    Simplified illustrations of \emph{discordance} and \emph{separability}.
    We assume that all sets are of size one, e.g., $\trnEmb = \{ \trnPoint \}$.
    Blue is better than red in (b) and (c).
    To minimize $\alignLoss = \setDist(\augPoint, \smash{\testPointA})$ as in (a), we propose the (b) discordance $\disVar$, which is the distance between $\smash{\testPointA}$ and the line segment $\ell$, and the (c) separability $\sepVar$, which is the distance between $\trnPoint$ and $\smash{\testPointA}$ projected onto $\ell$.
}
\label{figure:overview}
\vspaceBelowLargeFigureCaption
\end{figure}

\subsection{Main Ideas: Discordance and Separability}

Let $\genFunc \in \mathbb{R}^m \rightarrow \mathbb{R}^m$ be the underlying (unknown) anomaly-generating function in $\testData$, which transforms a normal data into an anomaly.
We aim to find $\augFunc$ that maximizes the functional similarity between $\augFunc$ and $\genFunc$, which we refer to \emph{alignment} in this work.
There are various ways to measure the alignment, but we focus on the embedding space, as it allows us to avoid the high dimensionality of real-world data.
We informally define the extent of alignment as follows.


\begin{proposition}
    Data augmentation function $\augFunc$ is aligned with the anomaly-generating function $\genFunc$ if $\alignLoss = \setDist(\augEmb, \smash{\testEmbA})$ is small.
\label{prop:alignment}
\end{proposition}

The problem is $\alignLoss$ cannot be computed without test labels.
To extract $\testEmbA$ from $\testEmb$ is as difficult as solving the anomaly detection problem itself.
Then, \emph{how can we approximate $\alignLoss$ {\textit{without}} test labels?}
We propose to decompose the alignment geometrically into \emph{discordance} $\disVar$ and \emph{separability} $\sepVar$ as shown in Fig. \ref{figure:overview}.
For an intuitive illustration, we assume that only one data exists in each set, e.g., $\trnEmb = \{ \trnPoint \}$.
Then, the simplified definitions of $\disVar$ and $\sepVar$ are given as
\begin{align}
    &\disVar(\trnPoint, \augPoint, \testPointA) = \frac{
        \setDist(\trnPoint, \testPointA) + \setDist(\augPoint, \testPointA)
    }{
        \setDist(\trnPoint, \augPoint)
    } - 1 \label{eq:dis-simple} \\
    &\sepVar(\trnPoint, \augPoint, \testPointA) = \frac{
        \mathrm{proj}(\trnPoint, \augPoint, \testPointA)
    }{
        \setDist(\trnPoint, \augPoint)
    }\;. \label{eq:sep-simple}
\end{align}


In combination, $\disVar$ and $\sepVar$ allow us to minimize $\alignLoss = \smash{d(\augPoint, \testPointA)}$ without actually computing it.
Let $\ell = \trnPoint + t(\augPoint - \trnPoint)$ be a line segment between $\trnPoint$ and $\augPoint$, where $t$ ranges over $[0, 1]$.
Then, $\disVar$ represents a distance between $\smash{\testPointA}$ and $\ell$, which is minimized when $\smash{\testPointA}$ is exactly on $\ell$.
On the other hand, $\sepVar$ means the distance between $\smash{\testPointA}$ and $\trnPoint$ when $\smash{\testPointA}$ is projected onto $\ell$.
Thus, $\alignLoss$ is minimized as zero if $\disVar = 0$ and $\sepVar = 1$.

A difference between $\disVar$ and $\sepVar$ is that $\disVar$ becomes a more accurate approximation of $\alignLoss$ if $\smash{\testPointA}$ is far from both $\trnPoint$ and $\augPoint$.
Thus, we consider $\disVar$ as a coarse-grained measure, while we bound the range of $\sepVar$ into $[0, 1]$ and consider it as a fine-grained measure to address the incapability of $\disVar$ to locate $\smash{\testPointA}$ on $\ell$.
Then, $\disVar$ is lower the better (alignment), while $\sepVar$ is higher the better.






The exact definitions of $\disVar$ and $\sepVar$ are direct generalization of Eq. \eqref{eq:dis-simple} and \eqref{eq:sep-simple} from vectors to sets.
The idea is to compute the average of all possible distances by replacing the vector distance with the set distance in Definition \ref{def:set-distance}:
\begin{align}
    &\disVar(\trnEmb, \augEmb, \testEmbA) = \frac{\setDist(\trnEmb, \testEmbA) + \setDist(\augEmb, \testEmbA)}{\setDist(\trnEmb, \augEmb)} - 1 \label{eq:dis-var} \\
    &\sepVar(\trnEmb, \augEmb, \testEmbA) = \frac{
        \sum_{\trnPoint, \augPoint, \testPointA \in \trnEmb, \augEmb, \smash{\testEmbA}}
        \mathrm{proj}(\trnPoint, \augPoint, \testPointA)
    }{
        \setDist(\trnEmb, \augEmb) |\trnEmb||\augEmb||\testEmbA|
    }. \label{eq:sep-var}
\end{align}

\myParagraph{Surrogate losses}
Based on our decomposition of the alignment, we propose surrogate losses $\scaledLoss$ and $\covLoss$ to approximate $\disVar$ and $\sepVar$, respectively, which have the term $\smash{\testEmbA}$ (unknown at test time) in their definitions.
Our final validation loss $\finalLoss$ is given as
\begin{equation}
\label{eq:valloss}
    \finalLoss(\trnEmb, \augEmb, \testEmb) = \scaledLoss(\cdot) - \frac{\max(\covLoss(\cdot), 1/2)}{\scaledLoss(\cdot)},
\end{equation}
where $\trnEmb$, $\augEmb$, and $\testEmb$ are inputs also to the right-hand side terms.
The minus sign is used since higher $\covLoss$ means better alignment until it reaches the optimum, which is $1/2$ in $\covLoss$, while it is $1$ for $\sepVar$.
We divide $\covLoss$ by $\scaledLoss$, since we want $\covLoss$ to have an effect especially when $\covLoss$ is small.
Then, we use $\finalLoss$ to perform unsupervised model selection by choosing the hyperparameters of $\augFunc$ that yields the smallest $\finalLoss$, which indicates the model with best alignment.




\subsection{Discordance Surrogate Loss}

We now describe how our surrogate losses $\scaledLoss$ and $\covLoss$ effectively approximate the discordance $\disVar$ and separability $\sepVar$, respectively.
$\scaledLoss$ is defined as
\begin{equation}
    \scaledLoss(\trnEmb, \augEmb, \testEmb) = 
    \frac{\setDist(\trnEmb \cup \augEmb, \testEmb)}{\setDist(\trnEmb, \augEmb)}.
\end{equation}

The idea is that $\setDist(\trnEmb \cup \augEmb, \testEmb)$ can approximate $\disVar$ based on the triangle inequality.
To show the exact relation between $\scaledLoss$ and $\disVar$, we first derive the lower and upper bounds of $\scaledLoss$ with respect to $\disVar$ in Lemma~\ref{theorem:scaled-property}.
Then, we show in Corollary~\ref{cor:scaled-linearity} that $\scaledLoss$ is represented as a linear function of $\disVar$ if some constraints are met, which makes $\scaledLoss$ an effective approximation of $\disVar$.


\begin{lemma}
    If $|\trnEmb| = |\augEmb|$, then the lower and upper bounds of $\scaledLoss$ are given as functions of $\disVar$ and $\setDist(\trnEmb, \augEmb)$:
    \begin{align*}
        c_2 \disVar + c_2 + c_3
        \leq \scaledLoss(\cdot)
        \leq c_2 \disVar + c_2 + c_3 + \frac{(c_1 + c_3)(\sigma + \epsilon)}{\setDist(\trnEmb, \augEmb)},
    \end{align*}
    where $c_i = \hat{c}_i / \sum_{k=1}^4 \hat{c}_k$ are data size-based constants  such that $\hat{c}_1 = |\trnEmb| \cdot |\smash{\testEmbN}|$, $\hat{c}_2 = |\trnEmb| \cdot |\smash{\testEmbA}|$, $\hat{c}_3 = |\augEmb| \cdot |\smash{\testEmbN}|$, and $\hat{c}_4 = |\augEmb|  \cdot|\smash{\testEmbA}|$.
\label{theorem:scaled-property}
\end{lemma}


\begin{proof}
    The proof is in Appendix \ref{appendix:proof-1}. \qed
\end{proof}

\begin{corollary}
    If $|\trnEmb| = |\augEmb|$, $\sigma \ll \setDist(\trnEmb, \augEmb)$, and $\epsilon \ll \setDist(\trnEmb, \augEmb)$, then $\scaledLoss$ is a linear function of $\disVar$: $\scaledLoss(\trnEmb, \augEmb, \testEmb) \approx c_2 h_d + c_2 + c_3$.
\label{cor:scaled-linearity}
\end{corollary}

\subsection{Separability Surrogate Loss}

The separability surrogate loss $\covLoss$ for approximating $\sepVar$ is defined as follows:
\begin{equation}
    \covLoss(\cdot)
    = \frac{
        \mathrm{std}(\{
            \mathrm{proj}(\mu_\mathrm{trn}, \mathbf{z}_\mathrm{aug}, \mathbf{z}_\mathrm{test})
            \mid
            \mathbf{z}_\mathrm{aug}, \mathbf{z}_\mathrm{test} \in \augEmb, \testEmb
        \})
    }{
        \setDist(\trnEmb, \augEmb)
    },
\end{equation}
where $\mathrm{std}(\mathcal{A}) = \sqrt{|\mathcal{A}|^{-1} \sum_{a \in \mathcal{A}} (a - \mathrm{mean}(\mathcal{A}))}$ is the standard variation of a set, and $\mu_\mathrm{trn}$ is the mean vector of $\trnEmb$.
One notable difference from Eq. \eqref{eq:sep-var} is that only the mean $\mu_\mathrm{trn}$ is used in the numerator, instead of whole $\trnEmb$, based on the observation that $\trnEmb$ is usually densely clustered as a result of training.


Intuitively, $\covLoss$ measures how much $\testEmb$ is scattered along the direction of $\mathbf{z}_\mathrm{aug} - \mu_\mathrm{trn}$.
The amount of scatteredness is directly related to the value of $\sepVar$, since we assume by convention that $\smash{\testEmbN}$ is close to $\trnEmb$.
In Lemma~\ref{lemma:cov-linearity}, we show that $\covLoss$ is a linear function of $\sepVar$ if some constraints are met, and its optimum is $1/2$ in the ideal case, which corresponds to $\sepVar = 1$, if $\bar{\sigma}_\mathrm{test} \ll \| \mathbf{z}_\mathrm{aug} - \mathbf{z}_\mathrm{trn} \|$.

\begin{lemma}
    We assume that $\trnEmb = \{ \mathbf{z}_\mathrm{trn} \}$, $\augEmb = \{ \mathbf{z}_\mathrm{aug} \}$, and $\smash{\mathbf{z}_\mathrm{test}^{(n)}} = \mathbf{z}_\mathrm{trn}$ for all $\smash{\mathbf{z}_\mathrm{test}^{(n)}} \in \smash{\testEmbN}$.
    Let $\gamma = |\smash{\testEmbA}| / |\testEmb|$, and $\bar{\sigma}_\mathrm{test}$ be the standard deviation of the projected norms $\smash{\testEmb^{(p)}} = \{ \mathrm{proj}(\mathbf{z}_\mathrm{trn}, \mathbf{z}_\mathrm{aug}, \mathbf{z}) \mid \mathbf{z} \in \smash{\testEmbA} \}$.
    Then, the separability surrogate loss $\covLoss$ is rewritten as a function of $\sepVar$ as follows:
    \begin{align*}
        \covLoss(\trnEmb, \augEmb, \testEmb) = \sqrt{\gamma (1 - \gamma)} \sepVar +  \frac{\sqrt{\gamma} \bar{\sigma}_\mathrm{test}}{\| \mathbf{z}_\mathrm{aug} - \mathbf{z}_\mathrm{trn} \|}.
    \end{align*}
\vspace{-5mm plus 1mm minus 1mm}
\label{lemma:cov-linearity}
\end{lemma}

\begin{proof}
    The proof is in Appendix \ref{appendix:proof-2}. \qed
\end{proof}

\begin{table}[t]
    \setlength\tabcolsep{5pt}
    \centering
    \caption{
        Average AUC (top) and rank (bottom) across 21 different tasks in the two datasets.
        The best is in bold, and the second best is underlined.
        Our \method achieves the best in six, and the second-best in two out of the 8 cases.
    }
\resizebox{\textwidth}{!}{
\begin{tabular}{l|cc|ccc|ccc|c}
	\toprule
	$\augFunc$ & Avg. & Rand. & Base & MMD & STD & MC & SEL & HITS & \textbf{\method} \\
	\midrule
	CutOut & 0.739 & \underline{0.776} & 0.741 & 0.735 & 0.739 & 0.749 & 0.727 & 0.757 & \textbf{0.813} \\
	CutAvg & 0.739 & \textbf{0.817} & 0.721 & 0.692 & 0.745 & 0.751 & 0.744 & 0.742 & \underline{0.806} \\
	CutDiff & 0.743 & 0.711 & 0.739 & 0.730 & 0.744 & 0.747 & 0.741 & \underline{0.777} & \textbf{0.811} \\
	CutPaste & 0.788 & 0.841 & 0.694 & 0.756 & 0.818 & \underline{0.862} & 0.830 & 0.850 & \textbf{0.884} \\
	\bottomrule
	\toprule
	$\augFunc$ & Avg. & Rand. & Base & MMD & STD & MC & SEL & HITS & \textbf{\method} \\
	\midrule
	CutOut & 7.33 & 6.10 & 6.62 & 6.93 & 6.29 & 6.50 & 7.10 & \underline{5.43} & \textbf{3.79} \\
	CutAvg & 7.00 & \underline{5.02} & 7.64 & 8.36 & 5.52 & 5.48 & 5.98 & 5.60 & \textbf{4.19} \\
	CutDiff & 6.43 & 7.24 & 6.45 & 7.38 & 6.00 & \underline{5.64} & 6.24 & 6.21 & \textbf{3.60} \\
	CutPaste & 7.67 & 6.29 & 8.67 & 7.21 & 5.60 & \textbf{4.33} & 5.17 & 4.64 & \underline{4.57} \\
	\bottomrule
\end{tabular}
}
    \label{tab:overall-accuracy}
\vspaceBelowLargeTable
\end{table}

\section{Experiments}

We answer the following questions through experiments on real datasets:
\begin{enumerate}
    \item[Q1.] \textbf{Performance.}
        Are the models selected by \method better than those selected by baseline measures for unsupervised model selection?
        Is the improvement statistically significant across different tasks and datasets? 
    \item[Q2.] \textbf{Ablation study.}
        Are the two main components of \method for the discordance and separability, respectively, meaningful to performance? 
        How do they complement each other across different augmentation functions and tasks?
    \item[Q3.] \textbf{Case studies.}
        How does \method work on individual cases with respect to the distribution of embedding vectors or anomaly scores?
\end{enumerate}

\subsection{Experimental Settings}
\label{subsec:baselines}


\myParagraph{Datasets}
We include two datasets for anomaly detection in natural images: MVTec AD \cite{Bergmann19MVTecAD} and MPDD \cite{Jezek21MPDD}, which contain 21 different tasks in total.
MVTec AD mimics real-world industrial inspection scenarios and contains 15 different tasks: five unique textures and ten unique objects from different domains.
MPDD focuses on defect detection during painted metal parts fabrication and contains 6 different object types with a non-homogeneous background.
The evaluation is done by AUC (the area under the ROC curve) scores on test data.

\myParagraph{Detector models}
We use a classifier-based anomaly detector model used in a previous work \cite{Li21CutPaste}, which first learns data embeddings and then computes anomaly scores on the space.
The model structure is based on ResNet18 \cite{He16ResNet}.
All model hyperparameters are set to the default setting, except for the number of training updates, which we changed for MPDD since the model converged much faster due to the smaller data size; we set the number of updates to 10,000 in MVTec AD, while to 1,000 in MPDD.

\begin{figure}[t]
    \centering
    \includegraphics[width=0.7\textwidth]{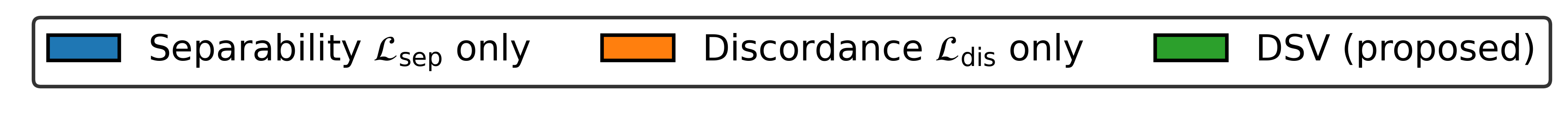}

    \vspace{-2mm}
    \includegraphics[width=\textwidth]{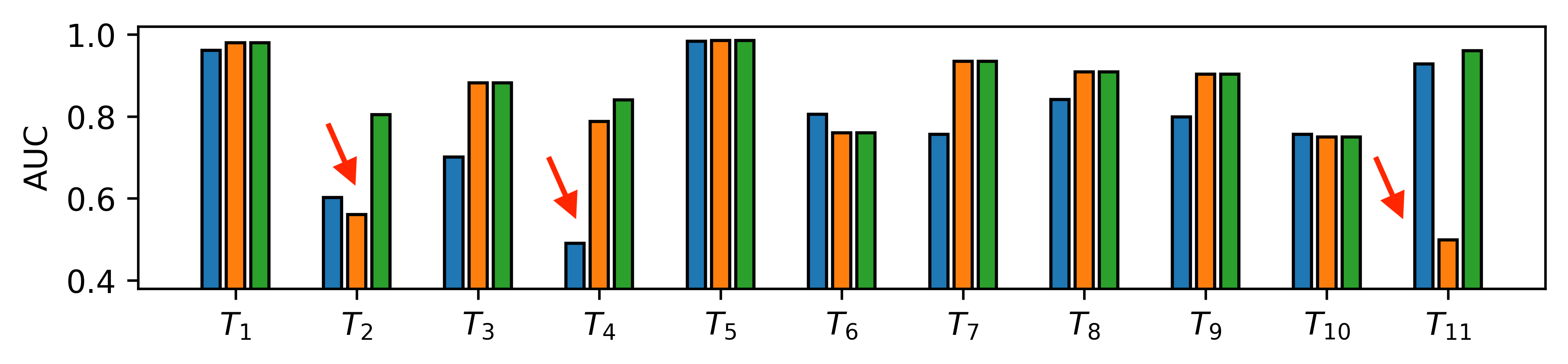}

    \vspace{-2mm}
    \includegraphics[width=\textwidth]{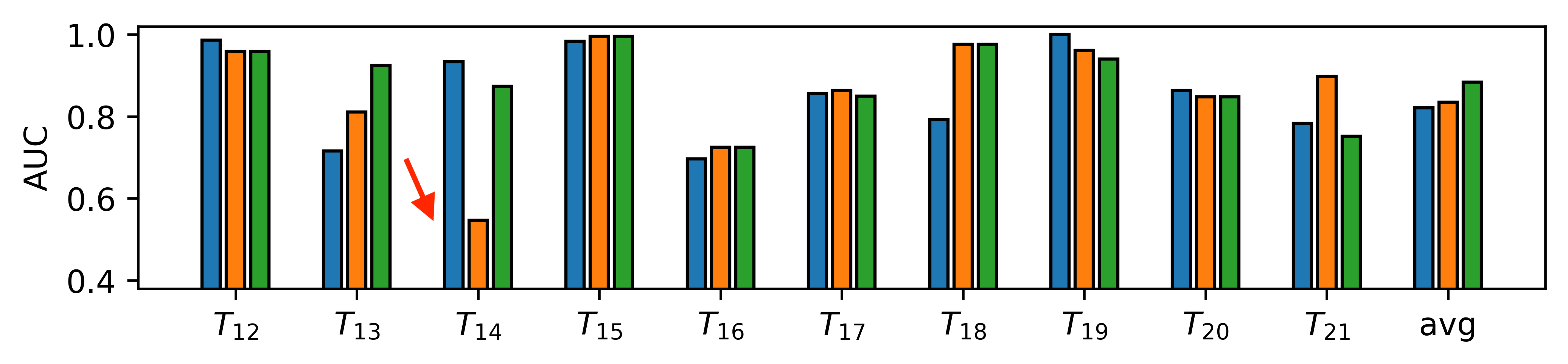}

    \vspace{-3mm}
    \caption{
        Ablation study to compare $\scaledLoss$, $\covLoss$, and $\finalLoss$ on 21 different tasks and on average when $\augFunc=\textrm{CutPaste}$.
        \method shows a dramatic improvement in a few cases, such as tasks $T_2$ (both fail), $T_4$ ($\covLoss$ fails),  $T_{11}$ and $T_{14}$ ($\scaledLoss$ fails).
    }
\vspaceBelowLargeFigureCaption
\label{fig:ablation}
\end{figure}

\myParagraph{Augmentation functions}
We use four different augmentation functions in experiments: CutOut \cite{Devries17CutOut}, CutAvg, CutDiff, and CutPaste \cite{Li21CutPaste}.
CutOut replaces a random patch from an image with black pixels.
CutAvg is similar to CutOut, but it replaces a patch with the average color of the patch, instead of the black.
CutDiff is a smooth version of CutOut, and it makes a smooth boundary when selecting a patch.
The resulting image has the black at the center of the original position of the patch, and it becomes brighter as it goes close to the boundary.
CutPaste copies a patch and pastes it into a random location of the image.

We use the patch size as the target augmentation hyperparameter to search for all these functions, since it directly controls the amount of modification by $\augFunc$.
We consider 17 settings in the range from $10^{-5}$ to $0.64$ in the log scale.
For example, $0.1$ represents we select a patch whose size is 10\% of the image. 


\myParagraph{Baselines}
We compare our \method with eight baseline methods for unsupervised model selection.
\emph{Average} is the simplest one, which is to take the average performance of all settings we consider.
\emph{Random} means we change the hyperparameter for each inference during training and test.
\emph{Base} is to use the distance $\scaledLoss(\trnEmb, \augEmb, \testEmb)$ as the simplest approximation of $\alignLoss$.
\emph{MMD} replaces the distance function in Base with the maximum mean discrepancy \cite{smola2006maximum}.
\emph{STD} measures standard deviation of the all-pair distances between $\trnEmb$ and $\testEmb$.

\begin{figure}[t]
    \centering
    \includegraphics[width=0.75\textwidth]{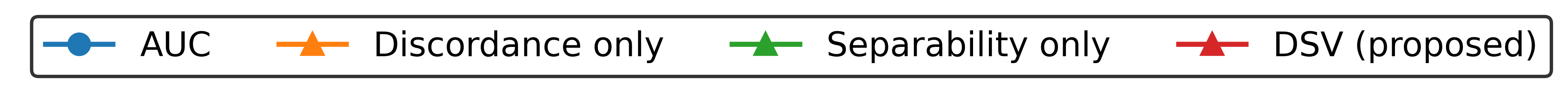}
    
    \begin{subfigure}{0.49\textwidth}
        \includegraphics[width=\textwidth]{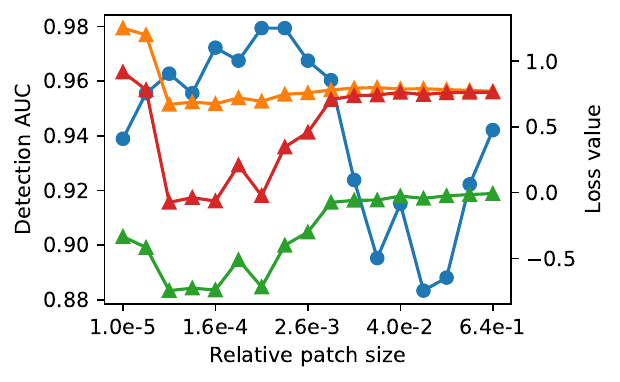}
        \caption{CutOut (Bottle)}
        \label{fig:auc-and-loss-1}
    \end{subfigure} \hfill
    \begin{subfigure}{0.49\textwidth}
        \includegraphics[width=\textwidth]{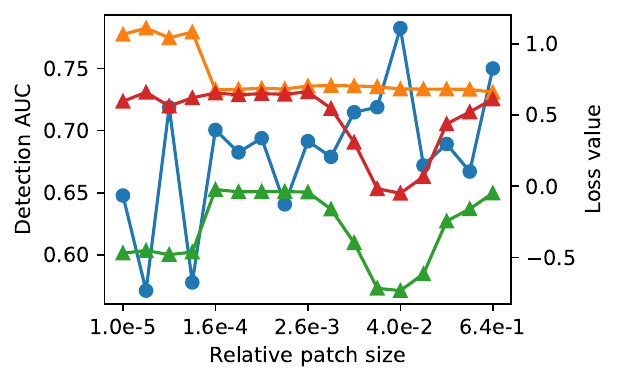}
        \caption{CutOut (Cable)}
        \label{fig:auc-and-loss-2}
    \end{subfigure}

    \vspace{1mm}

    \begin{subfigure}{0.49\textwidth}
        \includegraphics[width=\textwidth]{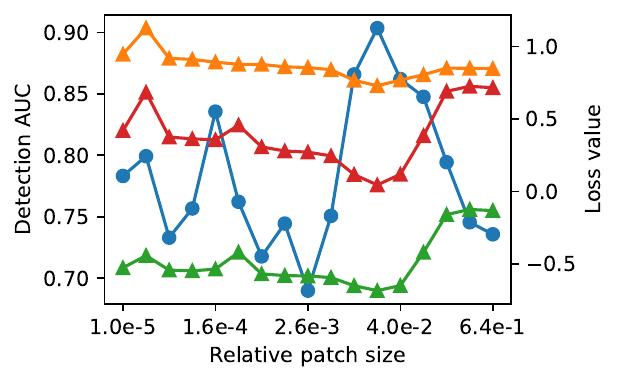}
        \caption{CutPaste (Pill)}
        \label{fig:auc-and-loss-3}
    \end{subfigure} \hfill
    \begin{subfigure}{0.49\textwidth}
        \includegraphics[width=\textwidth]{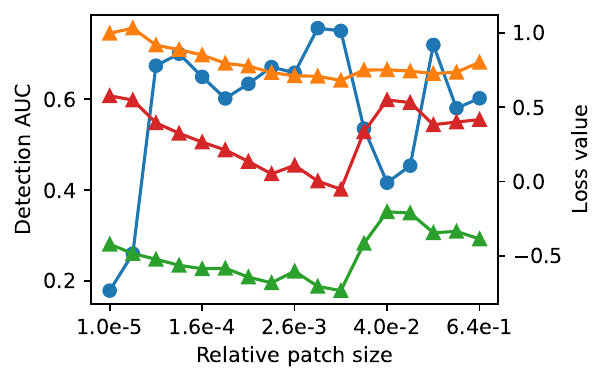}
        \caption{CutPaste (Screw)}
        \label{fig:auc-and-loss-4}
    \end{subfigure}

    \caption{
        The AUC and loss values $\scaledLoss$, $\covLoss$, and $\finalLoss$ with CutOut or CutPaste as $\augFunc$.
        We preprocessed $\covLoss$ so that it can be directly added to $\scaledLoss$ for creating $\finalLoss$.
        We have two main observations from the figures.
        First, $\finalLoss$ is negatively correlated with the actual AUC.
        Second, $\covLoss$ and $\scaledLoss$ work in a complementary way, which is shown especially well on (a) and (b).
    }
    \vspaceBelowLargeFigureCaption
\label{fig:auc-and-loss}
\end{figure}
 
MC, SEL, and HITS were proposed in a previous work \cite{ma2023need} for unsupervised outlier model selection (see \S \ref{subsec:uoms}). 
They are top-performing baselines based on internal performance measures.
MC \cite{Lin20InfoGAN,ma2023need} combines different models based on outlier score similarities, assuming that good models have similar outputs as the optimal model, and thus are close to each other.
HITS uses the HITS algorithm originally designed for web graphs \cite{Kleinberg99HITS} to compute the importance of each model.
SELECT (SEL in short) originates from model ensembles \cite{journals/sigkdd/ZimekCS13,rayana2016less}, and calculates the similarity between the output of each model and the ``pseudo ground truth'' which is initialized to the average of all candidate models.

\subsection{Detection Performance (Q1)}

\begin{wrapfigure}{r}{5.5cm}
    \vspace{-14mm}
    \includegraphics[width=5.5cm]{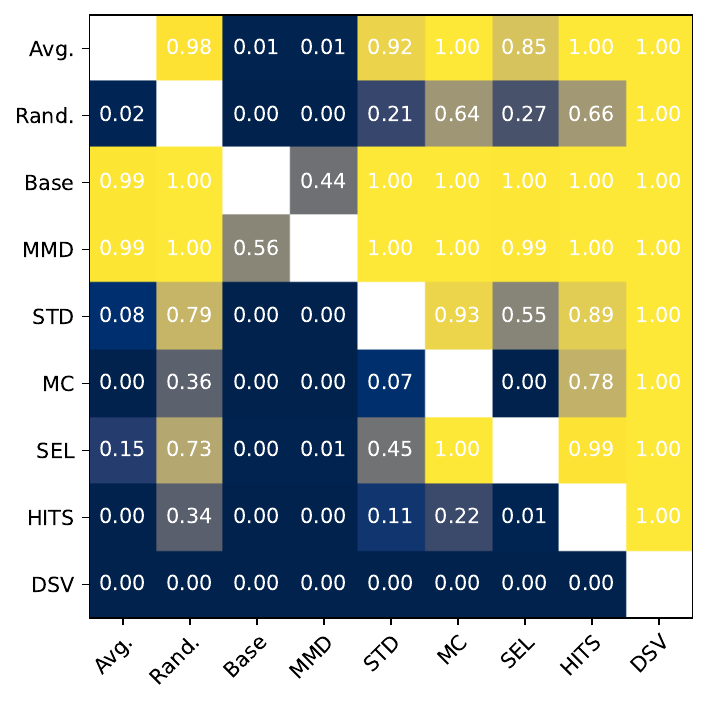}
    \caption{
        Wilcoxon signed rank test for all pairs of approaches.
        \method is superior to all other approaches with $p$-values smaller than $0.001$.
    }
    \vspace{-6mm}
\label{fig:wilcoxon}
\end{wrapfigure}

\begin{figure}[t]
    \centering
    \includegraphics[width=0.85\textwidth]{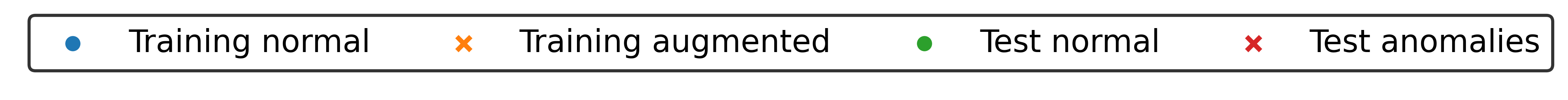}
    \vspace{-1mm}
    
    \begin{subfigure}{0.32\textwidth}
        \includegraphics[width=\textwidth]{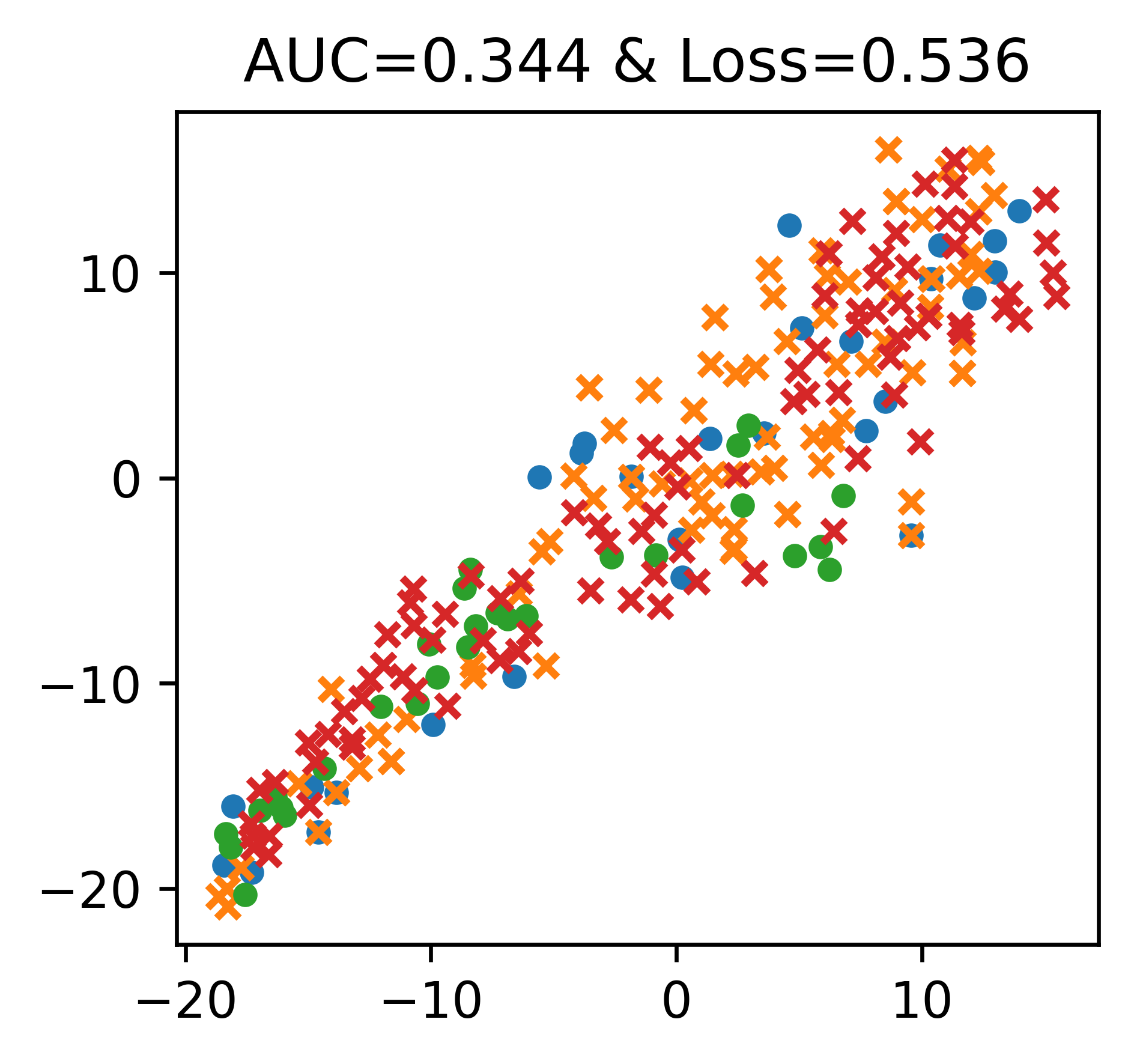}
        \caption{CutOut ($0.00004$)}
        \label{fig:embeddings-1}
    \end{subfigure} \hfill
    \begin{subfigure}{0.32\textwidth}
        \includegraphics[width=\textwidth]{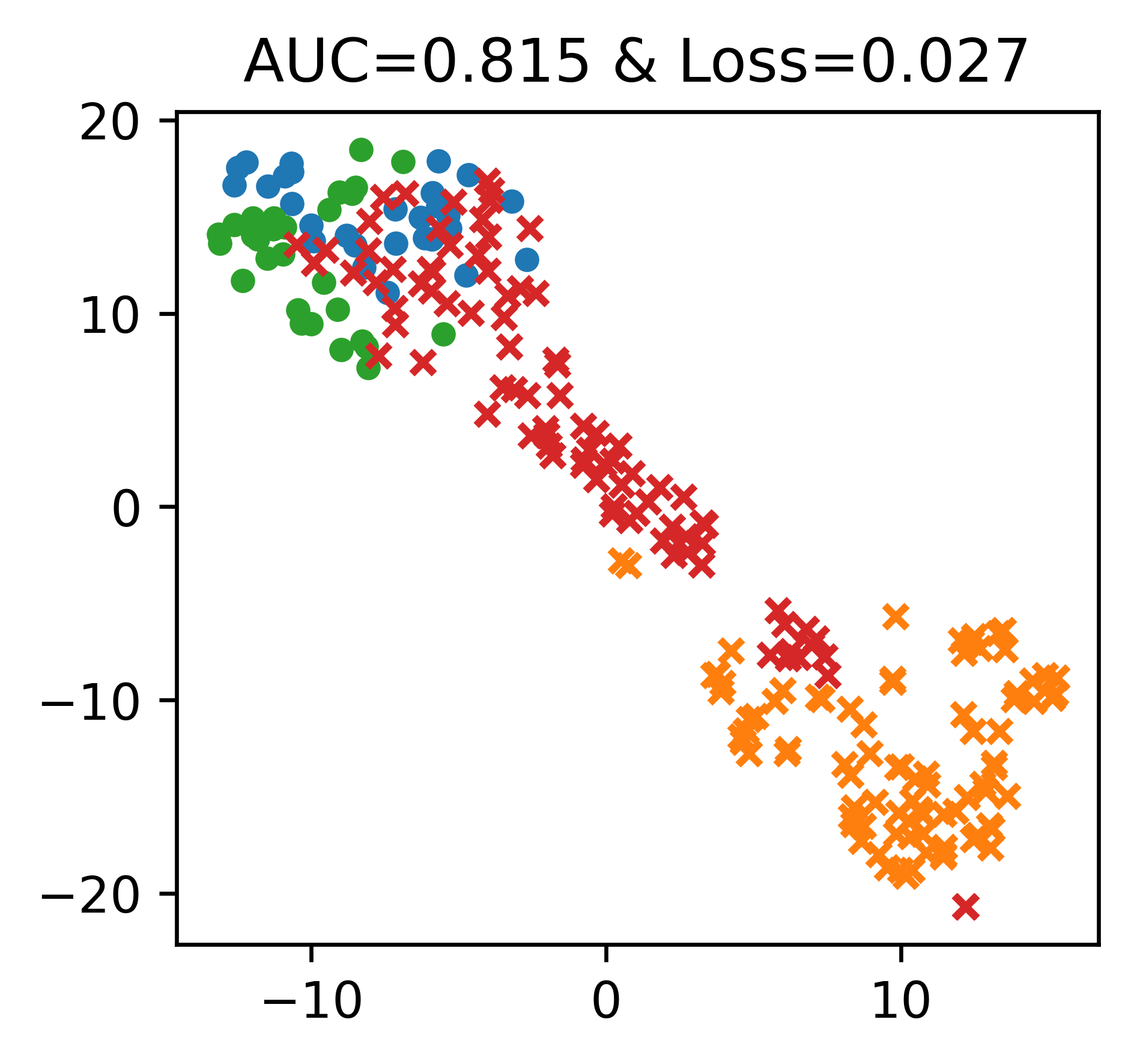}
        \caption{CutOut ($0.00128$)}
        \label{fig:embeddings-2}
    \end{subfigure} \hfill
    \begin{subfigure}{0.32\textwidth}
        \includegraphics[width=\textwidth]{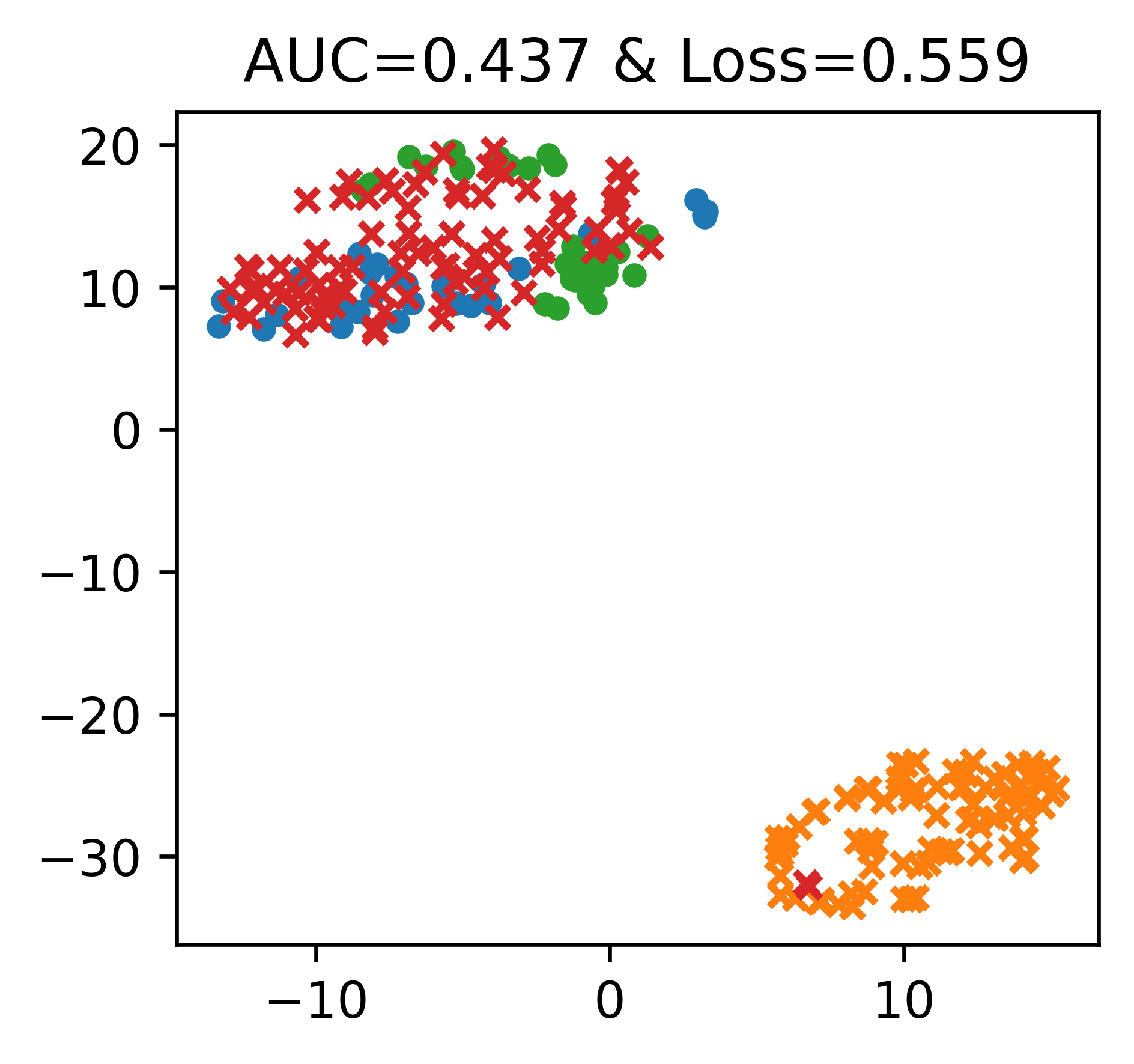}
        \caption{CutOut ($0.04$)}
        \label{fig:embeddings-3}
    \end{subfigure}

    \vspace{2mm}
    \begin{subfigure}{0.32\textwidth}
        \includegraphics[width=\textwidth]{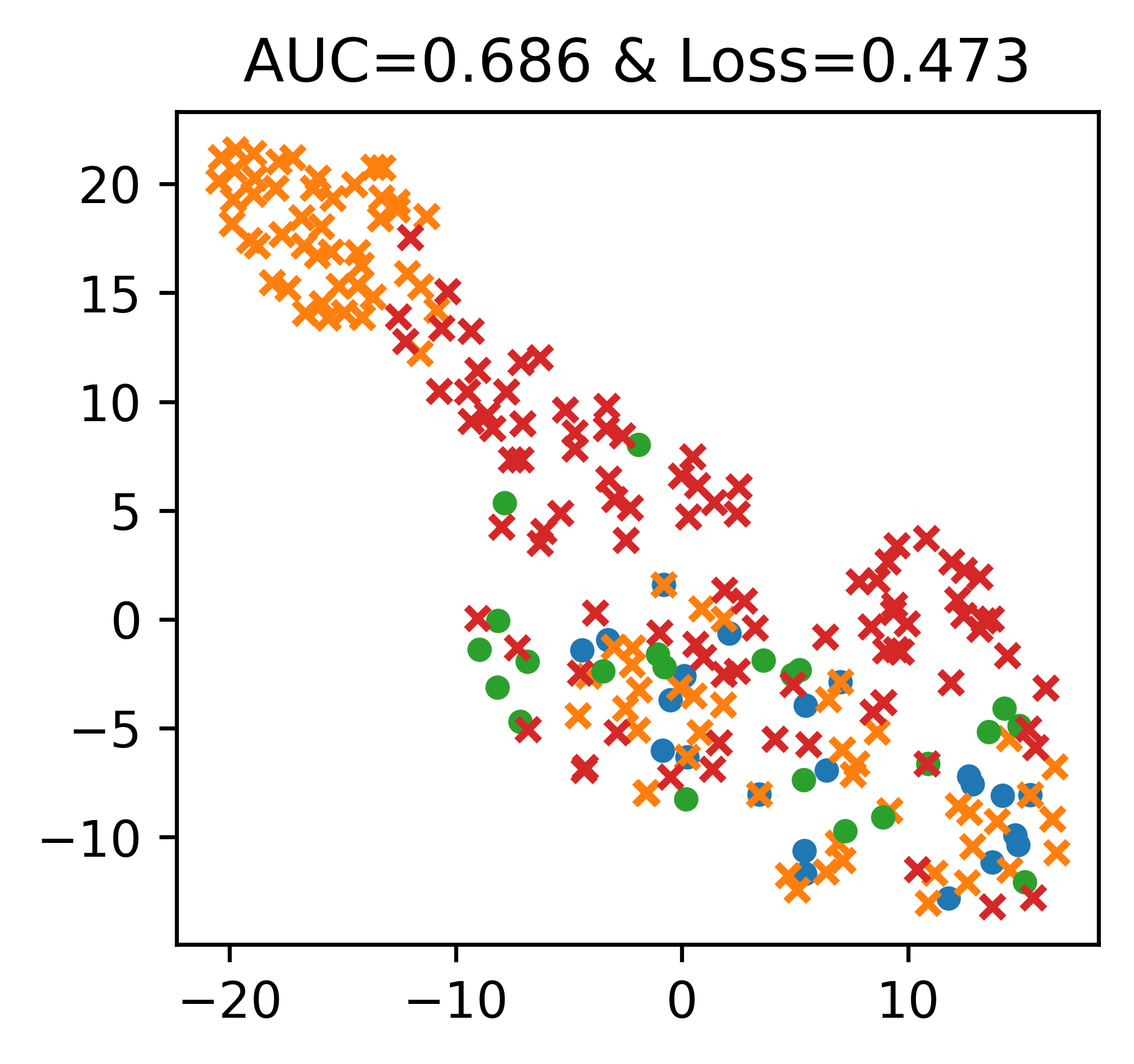}
        \caption{CutPaste ($0.00016$)}
        \label{fig:embeddings-4}
    \end{subfigure} \hfill
    \begin{subfigure}{0.32\textwidth}
        \includegraphics[width=\textwidth]{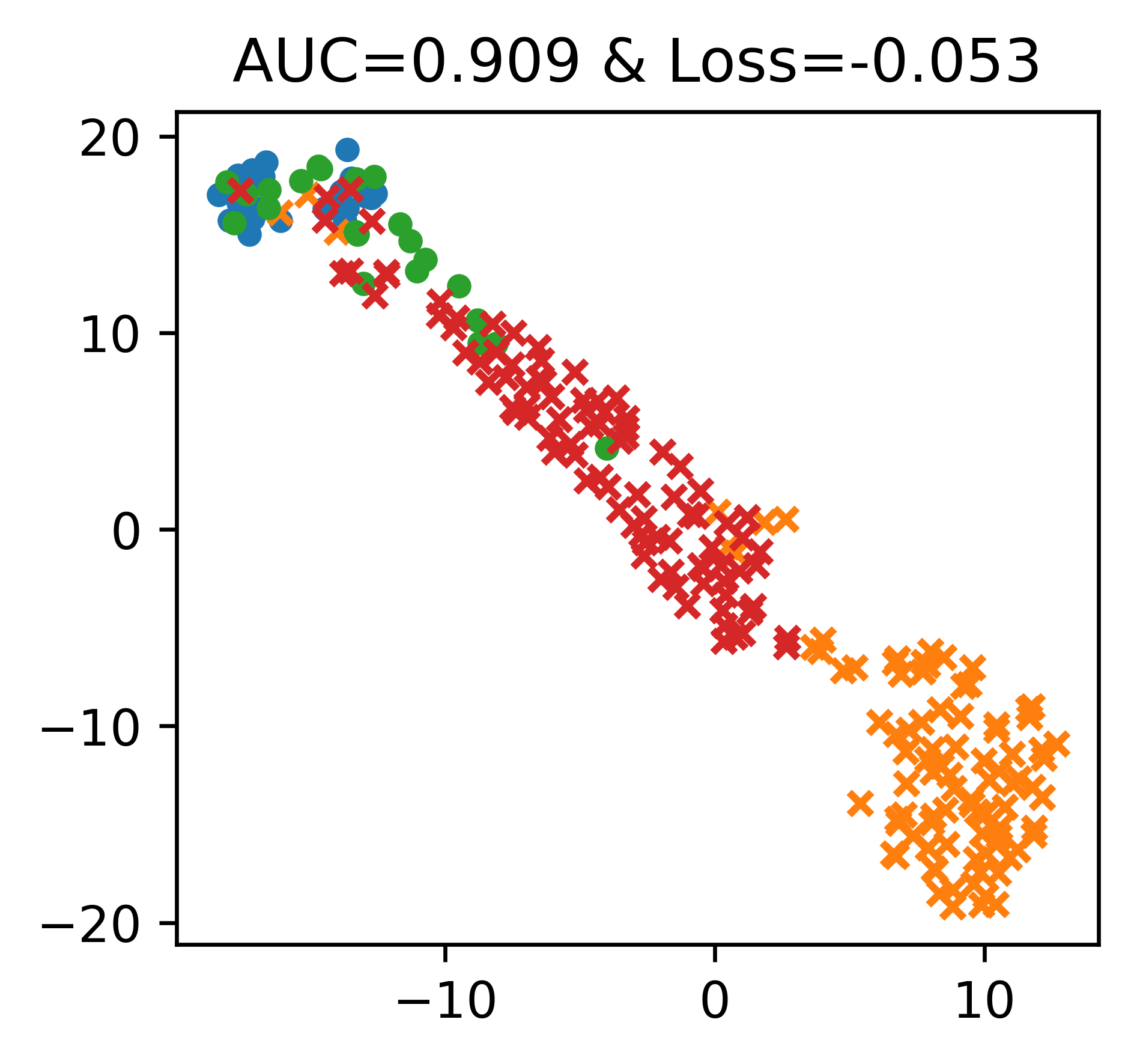}
        \caption{CutPaste ($0.02$)}
        \label{fig:embeddings-5}
    \end{subfigure} \hfill
    \begin{subfigure}{0.32\textwidth}
        \includegraphics[width=\textwidth]{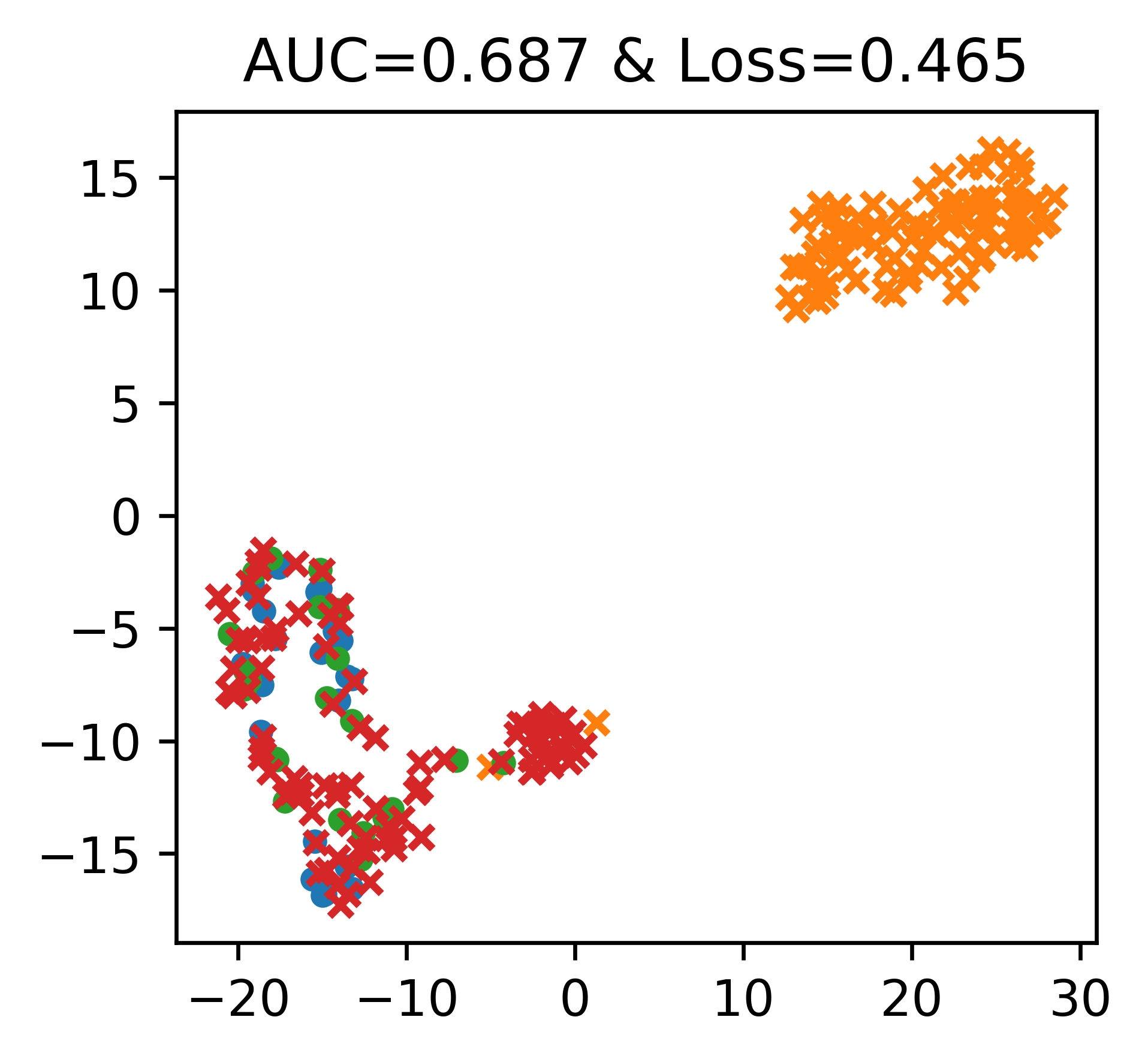}
        \caption{CutPaste ($0.32$)}
        \label{fig:embeddings-6}
    \end{subfigure}

    \caption{
        $t$-SNE visualizations of embeddings in (top) $\augFunc = \textrm{CutOut}$ and (bottom) $\augFunc = \textrm{CutPaste}$, where  values in parentheses represent different HPs.
        $\finalLoss$ is the smallest in (b) and (e), where the anomalies are in between $\trnEmb$ and $\augEmb$.
        Detection fails in (a), (c) 
        \& (d), (f), showing larger $\finalLoss$ than in (b) \& (e), resp.
    }
\label{fig:embeddings}
\vspaceBelowLargeFigureCaption
\end{figure}

Table \ref{tab:overall-accuracy} shows the average AUC and rank of various methods on 21 different tasks.
Due to the lack of space, we include the full results on individual tasks in the supplementary material.
\method shows the best performance on 6 out of the 8 cases, and the second-best on the remaining two cases.
MC and HITS perform well compared to the other baselines, but their performances are not consistent across different augmentation functions and tasks.

In Fig. \ref{fig:wilcoxon}, we perform the Wilcoxon signed rank test \cite{Groggel00Stats} to check if the differences between models are statistically significant.
Each number in the $(i, j)$-th cell represents the $p$-value comparing models $i$ and $j$, and it represents model $i$ is significantly better than model $j$ if the $p$-value is smaller than $0.05$.
\method is significantly better than all of the other approaches in the figure, demonstrating its superiority in unsupervised model selection.

\begin{figure}[t]
    \centering
    
    \begin{subfigure}{0.32\textwidth}
        \includegraphics[width=\textwidth]{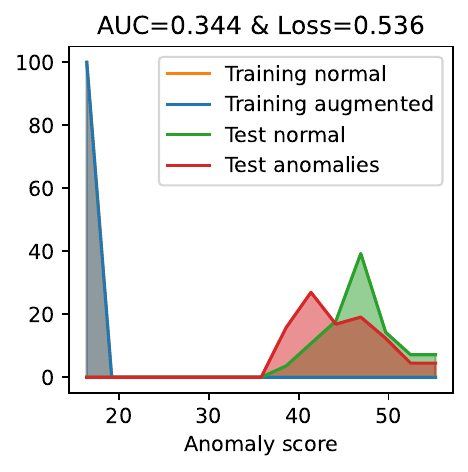}
        \caption{CutOut ($0.00004$)}
    \end{subfigure} \hfill
    \begin{subfigure}{0.32\textwidth}
        \includegraphics[width=\textwidth]{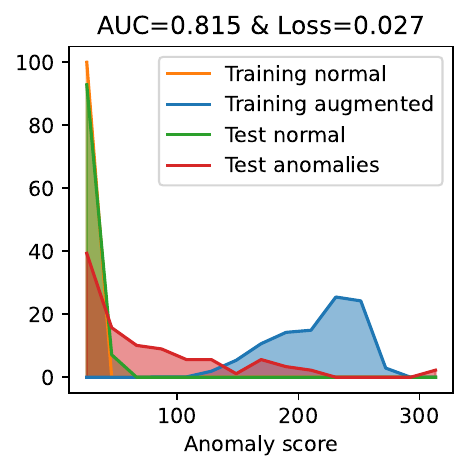}
        \caption{CutOut ($0.00128$)}
    \end{subfigure} \hfill
    \begin{subfigure}{0.32\textwidth}
        \includegraphics[width=\textwidth]{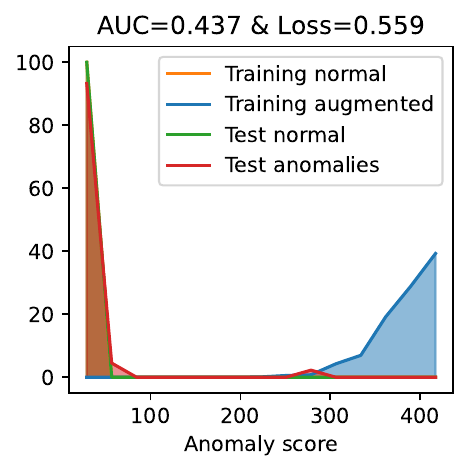}
        \caption{CutOut ($0.04$)}
    \end{subfigure}

    \caption{
        Anomaly scores for the three different HPs of $\augFunc=\textrm{CutOut}$ in Fig. \ref{fig:embeddings}.
        The distributions of embeddings are clearly observed also in the scores: (a) No separation in test data, (b) reasonable separation with as high AUC as $0.815$, and (c) drastic separation between 
        augmented points and all other sets.
    }
\label{fig:scores}
\vspaceBelowLargeFigureCaption
\end{figure}

\subsection{Ablation Studies (Q2)}

We perform an ablation study in Fig. \ref{fig:ablation}, comparing $\finalLoss$ with its two surrogate losses $\scaledLoss$ and $\covLoss$ on $\augFunc=\textrm{CutPaste}$.
The difference between the three models is more significant in individual cases, rather than on average, as denoted by the red arrows in the figure.
This is because each of $\scaledLoss$ and $\covLoss$ is incomplete by its design.
For example, $\disVar$ surpasses $\sepVar$ on average, but it shows some dramatic failure cases as in $T_{11}$ and $T_{14}$.
Our proposed $\finalLoss$ avoids such failures, achieving the best performance by effectively combining the two terms.

The complementary roles of the two losses is also shown in Fig. \ref{fig:auc-and-loss}, where we draw actual AUC and three different losses together for various combinations of $\augFunc$ and tasks.
Overall, the value of $\finalLoss$ is negatively correlated with the true AUC, which is exactly the purpose of introducing $\finalLoss$ for unsupervised model selection.
In detail, we observe complementary interactions between $\scaledLoss$ and $\covLoss$ from the figures; for example, in Fig. \ref{fig:auc-and-loss-1}, $\covLoss$ makes the overall loss decrease when AUC peaks the top, although $\scaledLoss$ makes only negligible changes.
In Fig. \ref{fig:auc-and-loss-2}, in contrast, the two losses change drastically in small patch sizes, while their sums remain similar, allowing us to avoid HPs with low AUC.

\subsection{Case Studies (Q3)}

In Fig. \ref{fig:embeddings}, we visualize the embeddings when $\augFunc = \textrm{CutOut}$ (the task is Carpet) and $\augFunc = \textrm{CutPaste}$ (the task is Metal Nut).
In Figs. \ref{fig:embeddings-2} and \ref{fig:embeddings-5}, which show the smallest $\finalLoss$, test anomalies $\smash{\testEmbA}$ are scattered in between $\trnEmb$ and $\augEmb$.
Although some of $\smash{\testEmbA}$ are mixed with $\smash{\testEmbN}$ in Fig. \ref{fig:embeddings-2}, the AUC is as high as $0.815$.
On the other hand, in Figs. \ref{fig:embeddings-1} and \ref{fig:embeddings-3}, the AUC is lower than even $0.5$, while $\finalLoss$ is large.
In Fig. \ref{fig:embeddings-1}, $\smash{\testEmbN}$ and $\smash{\testEmbA}$ are mixed completely, since the amount of modification through augmentation is too small.
In Fig. \ref{fig:embeddings-3}, 
$\augEmb$ are separated from all other sets, due to the drastic augmentation.
Fig.s \ref{fig:embeddings-4} and \ref{fig:embeddings-6} show similar patterns, although the AUC is generally higher than in CutOut.

In Fig. \ref{fig:scores}, we visualize the anomaly scores generated by our detector model, following the same scenarios as in Fig. \ref{fig:embeddings} when $\augFunc=\textrm{CutOut}$.
Since the detector model in our experiments computes an anomaly score based on the likelihood of a Gaussian mixture model in the embedding space, the scores are related to the actual distances.
The scores represent the difference between different HPs well, leading to the observations consistent with the $t$-SNE visualization.

\section{Conclusion}

There has been a recent surge of self-supervised learning methods for anomaly detection (SSAD), but how to systematically choose the augmentation hyperparameters here remains vastly understudied.
To address this, we introduce \method, an unsupervised validation loss for selecting optimal SSAD models with effective augmentation hyperparameters.
The main idea is to maximize the alignment between augmentation and unknown anomalies with surrogate losses that estimate the discordance and separability of test data.
Our experiments demonstrate that \method outperforms a broad range of baselines.
Future work involves extending it to incorporate other distance measures such as the Chebyshev distance.

\subsubsection{Acknowledgements}
This work is partially sponsored by PwC Risk and Regulatory Services Innovation Center at Carnegie Mellon University. Any conclusions expressed in this material are those of the author and do not necessarily reflect the views, expressed or implied, of the funding parties.

\appendix
\section{Proofs of Lemmas}



\subsection{Proof of Lemma \ref{theorem:scaled-property}}
\label{appendix:proof-1}

\begin{proof}
Let $\hat{\sigma} = \sigma + \epsilon$.
We rewrite the numerator of $\scaledLoss$ based on the definition of $h$ and Assumption \ref{assume:test-data}.
\begin{multline*}
    \setDist(\trnEmb \cup \augEmb, \testEmb)
        = c_1 \hat{\sigma}
        + c_2 ((1 + h) \setDist(\trnEmb, \augEmb) - \setDist(\augEmb, \testEmbA)) \\
        + c_3 \setDist(\augEmb, \testEmbN)
        + c_4 \setDist(\augEmb, \testEmbA)
\end{multline*}


Then, we derive the lower bound as follows:
\begin{align*}
    &\setDist(\trnEmb \cup \augEmb, \testEmb) \\
    &\quad \quad \geq
        c_2 ((1 + h) \setDist(\trnEmb, \augEmb) - \setDist(\augEmb, \testEmbA))
        \\ &\quad \quad \quad
        + c_3 \setDist(\trnEmb, \augEmb)
        + c_4 \setDist(\augEmb, \testEmbA)
        + (c_1 - c_3) \hat{\sigma}
        \\
    &\quad \quad =
        (c_4 - c_2) \setDist(\augEmb, \testEmbA)
        + (c_2 + c_2h + c_3) \setDist(\trnEmb, \augEmb)
        + (c_1 - c_3) \hat{\sigma}
\end{align*}


Similarly, the upper bound is given as follows: 
\begin{align*}
    &\setDist(\trnEmb \cup \augEmb, \testEmb) \\
    &\quad \quad \leq
        c_2 ((1 + h) \setDist(\trnEmb, \augEmb) - \setDist(\augEmb, \testEmbA))
        \\ &\quad \quad \quad
        + c_3 \setDist(\trnEmb, \augEmb)
        + c_4 \setDist(\augEmb, \testEmbA)
        + (c_1 + c_3) \hat{\sigma}
        \\
    &\quad \quad =
        (c_4 - c_2) \setDist(\augEmb, \testEmbA)
        + (c_2 + c_2h + c_3) \setDist(\trnEmb, \augEmb)
        + (c_1 + c_3) \hat{\sigma}
\end{align*}

If we apply the assumption $|\trnEmb| = |\augEmb|$, which results in $c_2 = c_4$, the first term from both bounds disappears.
We get the inequalities in the lemma by dividing both bounds by $\setDist(\trnEmb, \augEmb)$.
\qed
\end{proof}

\subsection{Proof of Lemma \ref{lemma:cov-linearity}}
\label{appendix:proof-2}

\begin{proof}
Let $\mu_\mathrm{test} = \mathrm{mean}(\{ \mathrm{proj}(\mathbf{z}_\mathrm{trn}, \mathbf{z}_\mathrm{aug}, \mathbf{z}) \mid \mathbf{z} \in \testEmbA \} )$ be the average of projected norms.
We first rewrite $\sepVar$ as follows:
\begin{align*}
    \sepVar
    &= \frac{
        \sum_{\trnPoint, \augPoint, \testPointA \in \trnEmb, \augEmb, \smash{\testEmbA}}
        \mathrm{proj}(\trnPoint, \augPoint, \testPointA)
    }{
        \setDist(\trnEmb, \augEmb) |\trnEmb||\augEmb||\testEmbA|
    }
    \\ &=
    \frac{
        \sum_{\testPointA \in \testEmbA}
        \mathrm{proj}(\trnPoint, \augPoint, \testPointA)
    }{
        \| \augPoint - \trnPoint \| |\testEmbA|
    }
    \\ &=
    \frac{
        |\testEmbA| \mu_\mathrm{test}
    }{
        \| \augPoint - \trnPoint \| |\testEmbA|
    }
    = 
    \frac{
        \mu_\mathrm{test}
    }{
        \| \augPoint - \trnPoint \|
    }
\end{align*}

We rewrite the \emph{squared} numerator of $\covLoss$:
\begin{align*}
    &\mathrm{std}^2(\{
        \mathrm{proj}(\mu_\mathrm{trn}, \mathbf{z}_\mathrm{aug}, \mathbf{z}_\mathrm{test})
        \mid
        \mathbf{z}_\mathrm{aug}, \mathbf{z}_\mathrm{test} \in \augEmb, \testEmb
    \})
    \\ &\quad \quad =
    \mathrm{std}^2(\{
        \mathrm{proj}(\mu_\mathrm{trn}, \mathbf{z}_\mathrm{aug}, \mathbf{z}_\mathrm{test})
        \mid
        \mathbf{z}_\mathrm{test} \in \testEmb
    \})
    \\ &\quad \quad =
    \frac{1}{|\testEmb|} \sum_{\testPoint}(
        \mathrm{proj}(\mu_\mathrm{trn}, \mathbf{z}_\mathrm{aug}, \mathbf{z}_\mathrm{test})
        - \gamma \mu_\mathrm{test}
    )^2
    \\ &\quad \quad =
    \frac{1}{|\testEmb|} \left(
        |\testEmbN| \gamma^2 \mu_\mathrm{test}^2
        + |\testEmbA| (\bar{\sigma}_\mathrm{test}^2 + (1 - \gamma)^2 \mu_\mathrm{test}^2)
    \right)
    \\ &\quad \quad =
        (1 - \gamma) \gamma^2 \mu_\mathrm{test}^2
        + \gamma (\bar{\sigma}_\mathrm{test}^2 + (1 - \gamma)^2 \mu_\mathrm{test}^2)
    \\ &\quad \quad =
        \gamma (1 - \gamma) \mu_\mathrm{test}^2
        + \gamma \bar{\sigma}_\mathrm{test}^2.
\end{align*}

Then, $\covLoss$ is rewritten as follows:
\begin{align*}
    \covLoss
    &= \frac{\sqrt{\gamma (1 - \gamma) \mu_\mathrm{test}^2
        + \gamma \bar{\sigma}_\mathrm{test}^2}}{\setDist(\trnEmb, \augEmb)} 
    = \sqrt{\gamma (1 - \gamma)} \sepVar +  \frac{\sqrt{\gamma} \bar{\sigma}_\mathrm{test}}{\| \mathbf{z}_\mathrm{aug} - \mathbf{z}_\mathrm{trn} \|},
\end{align*}
which is the equation in the lemma.
\qed
\end{proof}

%
%
%
\newpage
\bibliographystyle{splncs04}
\bibliography{main}

\begin{thebibliography}{10}
\providecommand{\url}[1]{\texttt{#1}}
\providecommand{\urlprefix}{URL }
\providecommand{\doi}[1]{https://doi.org/#1}

\bibitem{baevski2022data2vec}
Baevski, A., Hsu, W.N., Xu, Q., Babu, A., Gu, J., Auli, M.: Data2vec: A general
  framework for self-supervised learning in speech, vision and language. In:
  International Conference on Machine Learning. pp. 1298--1312. PMLR (2022)

\bibitem{Bergman20GOAD}
Bergman, L., Hoshen, Y.: Classification-based anomaly detection for general
  data. In: ICLR (2020)

\bibitem{Bergmann19MVTecAD}
Bergmann, P., Fauser, M., Sattlegger, D., Steger, C.: Mvtec {AD} - {A}
  comprehensive real-world dataset for unsupervised anomaly detection. In: CVPR
  (2019)

\bibitem{chen2021empirical}
Chen, X., Xie, S., He, K.: An empirical study of training self-supervised
  vision transformers. In: ICCV (2021)

\bibitem{Devries17CutOut}
Devries, T., Taylor, G.W.: Improved regularization of convolutional neural
  networks with cutout. CoRR  \textbf{abs/1708.04552} (2017)

\bibitem{elnaggar2021prottrans}
Elnaggar, A., Heinzinger, M., Dallago, C., Rehawi, G., Wang, Y., Jones, L.,
  Gibbs, T., Feher, T., Angerer, C., Steinegger, M., et~al.: Prottrans: Toward
  understanding the language of life through self-supervised learning. IEEE
  transactions on pattern analysis and machine intelligence  \textbf{44}(10),
  7112--7127 (2021)

\bibitem{Golan18GEOM}
Golan, I., El{-}Yaniv, R.: Deep anomaly detection using geometric
  transformations. In: NeurIPS (2018)

\bibitem{Groggel00Stats}
Groggel, D.J.: Practical nonparametric statistics. Technometrics
  \textbf{42}(3),  317--318 (2000)

\bibitem{He16ResNet}
He, K., Zhang, X., Ren, S., Sun, J.: Deep residual learning for image
  recognition. In: CVPR (2016)

\bibitem{Jezek21MPDD}
Jezek, S., Jonak, M., Burget, R., Dvorak, P., Skotak, M.: Deep learning-based
  defect detection of metal parts: evaluating current methods in complex
  conditions. In: ICUMT (2021)

\bibitem{Kleinberg99HITS}
Kleinberg, J.M.: Authoritative sources in a hyperlinked environment. J. {ACM}
  \textbf{46}(5),  604--632 (1999)

\bibitem{Kolesnikov19Revisiting}
Kolesnikov, A., Zhai, X., Beyer, L.: Revisiting self-supervised visual
  representation learning. In: CVPR (2019)

\bibitem{Li21CutPaste}
Li, C., Sohn, K., Yoon, J., Pfister, T.: Cutpaste: Self-supervised learning for
  anomaly detection and localization. In: CVPR (2021)

\bibitem{Lin20InfoGAN}
Lin, Z., Thekumparampil, K.K., Fanti, G., Oh, S.: Infogan-cr and
  modelcentrality: Self-supervised model training and selection for
  disentangling gans. In: ICML (2020)

\bibitem{ma2023need}
Ma, M.Q., Zhao, Y., Zhang, X., Akoglu, L.: The need for unsupervised outlier
  model selection: A review and evaluation of internal evaluation strategies.
  ACM SIGKDD Explorations Newsletter  \textbf{25}(1) (2023)

\bibitem{mackay2019self}
MacKay, M., Vicol, P., Lorraine, J., Duvenaud, D., Grosse, R.: Self-tuning
  networks: Bilevel optimization of hyperparameters using structured
  best-response functions. arXiv preprint arXiv:1903.03088  (2019)

\bibitem{Qiu21NeuTraL}
Qiu, C., Pfrommer, T., Kloft, M., Mandt, S., Rudolph, M.: Neural transformation
  learning for deep anomaly detection beyond images. In: ICML (2021)

\bibitem{rayana2016less}
Rayana, S., Akoglu, L.: Less is more: Building selective anomaly ensembles. ACM
  Trans. Knowl. Discov. Data  \textbf{10}(4),  42:1--42:33 (2016)

\bibitem{Sehwag21SSD}
Sehwag, V., Chiang, M., Mittal, P.: {SSD:} {A} unified framework for
  self-supervised outlier detection. In: ICLR (2021)

\bibitem{Shenkar22ICL}
Shenkar, T., Wolf, L.: Anomaly detection for tabular data with internal
  contrastive learning. In: ICLR (2022)

\bibitem{smola2006maximum}
Smola, A.J., Gretton, A., Borgwardt, K.: Maximum mean discrepancy. In: 13th
  international conference, ICONIP. pp.~3--6 (2006)

\bibitem{Sohn21DeepOC}
Sohn, K., Li, C., Yoon, J., Jin, M., Pfister, T.: Learning and evaluating
  representations for deep one-class classification. In: ICLR (2021)

\bibitem{wolpert1997no}
Wolpert, D.H., Macready, W.G.: No free lunch theorems for optimization. IEEE
  transactions on evolutionary computation  \textbf{1}(1),  67--82 (1997)

\bibitem{Ye21Bias}
Ye, Z., Chen, Y., Zheng, H.: Understanding the effect of bias in deep anomaly
  detection. In: IJCAI (2021)

\bibitem{yoo2022role}
Yoo, J., Zhao, T., Akoglu, L.: Self-supervision is not magic: Understanding
  data augmentation in image anomaly detection. arXiv  (2022)

\bibitem{zhao2021automatic}
Zhao, Y., Rossi, R., Akoglu, L.: Automatic unsupervised outlier model
  selection. Advances in Neural Information Processing Systems  \textbf{34},
  4489--4502 (2021)

\bibitem{zhao2022elect}
Zhao, Y., Zhang, S., Akoglu, L.: Toward unsupervised outlier model selection.
  In: ICDM. pp. 773--782. {IEEE} (2022)

\bibitem{journals/sigkdd/ZimekCS13}
Zimek, A., Campello, R.J.G.B., Sander, J.: Ensembles for unsupervised outlier
  detection: challenges and research questions a position paper. SIGKDD Explor.
   \textbf{15}(1),  11--22 (2013)

\bibitem{zoph2020learning}
Zoph, B., Cubuk, E.D., Ghiasi, G., Lin, T.Y., Shlens, J., Le, Q.V.: Learning
  data augmentation strategies for object detection. In: ECCV (2020)

\end{thebibliography}


\end{document}